\documentclass{llncs}
\usepackage[T1]{fontenc}
\usepackage{todonotes} 
\usepackage[utf8]{inputenc}
\usepackage{pgfplotstable}
\usepackage{pgfplots}
\usepackage{times}
\usepackage{latexsym}
\usepackage{multirow}
\usepackage{booktabs}
\usepackage{amssymb}
\usepackage{graphicx}
\usepackage{fancyvrb}
\usepackage{times}
\usepackage{standalone}

\usepackage{tikz}
\usetikzlibrary{positioning, calc, arrows.meta, decorations.pathreplacing, pgfplots.groupplots}

\tikzset{
  docs/.style={
    draw, fill=white, minimum size=5mm
  }
}

\title{On the Influence of Machine Translation on Language Origin Obfuscation}
\author{Benjamin Murauer \and Michael Tschuggnall \and Günther Specht\\ {\tt\{firstname.lastname\}@uibk.ac.at}}
\institute{Universität Innsbruck}

\VerbatimFootnotes

\begin{document}
\maketitle
\begin{abstract}
  In the last decade, machine translation has become a popular means to deal with multilingual digital content.
  By providing higher quality translations, obfuscating the source language of a text becomes more attractive. 
  In this paper, we analyze the ability to detect the source language from the translated output of two widely used commercial machine translation systems by utilizing machine-learning algorithms with basic textual features like n-grams.
  Evaluations show that the source language can be reconstructed with high accuracy for documents that contain a sufficient amount of translated text.
  In addition, we analyze how the document size influences the performance of the prediction, as well as how limiting the set of possible source languages improves the classification accuracy.
\end{abstract}

\section{Introduction}
The amount of textual data being shared through various channels over the Internet is growing constantly.
Considering the fact that this data is transferred across cultural as well as political borders, machine-translation is an often used means in case the source or target language is understood insufficiently.
On the other hand, the growing quality of publicly available services from, e.g., Google or Microsoft, can also be used to obfuscate the source language, and thus the geographical origin.
While there exist many approaches in the field of stylometry to identify the author of documents \cite{Stamatatos2009} (which is also frequently used in digital forensics \cite{Nirkhi2013}), to determine the native language of a person based on samples written in a non-native language (Native Language Identification, e.g., \cite{Ionescu2014}), or to detect the translation tool of a machine-translated text \cite{Aharoni2014} it has not been investigated whether similar techniques can be utilized to predict the source language of machine-translated text with satisfying accuracy. Algorithms which are able to reconstruct the original language could be utilized for, e.g.,  detecting plagiarism: once a part of a document is confirmed to be machine-translated and the original language is known, further investigations can focus on that languages. Regarding forensics, a use case could include determining the native language of a threat that has been automatically translated, e.g., to restrict the geographical origin and consequently the amount of possible authors.

We analyze the prediction performance of commonly used stylometric features such as n-grams -- which are known to work well for various text classfication problems \cite{Ionescu2014,Malmasi2017} -- regarding the detection of the source language from machine-translated text.
Adhering to the fact that most of online text is shared in English \cite{webenglish}, we restrict the target language to be English as well for this study.

Analogous to the field of authorship attribution where it has been shown that an increasing amount of possible authors decreases the prediction performance, we additionally analyze the influence of the number of possible source languages on the quality of the prediction.
Finally, we determine the amount of text that is required to achieve a specific accuracy for predicting the source language. 

Concretely, we answer two main research questions (RQ) in this paper:
\begin{itemize}
  \item[] \textbf{RQ1}: Given a document, can it be detected whether it was machine-translated, and if yes, how accurate can the source language be determined?
  \item[] \textbf{RQ2}: How do different feature sets, varying amounts of possible source languages and document lengths influence the prediction accuracy?
\end{itemize}

We find that using traditional features, the original source language of a translated text can be predicted with a very high accuracy. 
Additionally, we show that this prediction is possible with even very small amounts of text for many possible languages.

\section{Related Work}

%\subsection{Authorship Attribution}
Previous work in authorship attribution \cite{Caliskan2012,Rao2000} has shown that the use of automatic translation techniques actually adds information to a text document since the translation itself adds details specific to the methods used.
We assume this fact is still valid although the engine used by the translation services has changed from statistical methods to neural networks\footnote{https://www.blog.google/products/translate/found-translation-more-accurate-fluent-sentences-google-translate/}\textsuperscript{,}\footnote{https://blogs.msdn.microsoft.com/translation/2016/11/15/microsoft-translator-launching-neural-network-based-translations-for-all-its-speech-languages/}.
%\cite{Microsoft2016,Turovsky2016}.

%\subsection{Native Language Identification}
A field of research closely related to the problem presented in this paper is Native Language Identification (NLI), which aims to find the native language of authors given only documents in a foreign language.
Features used for this classification task range from traditional n-grams \cite{Malmasi2014,Bykh2012} over string kernels \cite{Ionescu2017} to detailed stylistic analyses of the words and grammar that was used by the authors \cite{Bykh2016,Bykh2014,Swanson2013,Swanson2014}.

In an approach more closely related to the problem addressed by this paper, Koppel et al. determine the source language of human-translated documents by analyzing various types of errors introduced by non-native authors \cite{Koppel2005}.
%In \cite{Bykh2016}, Bykh et al. analyze specific variations of words that are chosen by authors, which in combination with various types of n-grams and detailed context free grammar production rules achieves an accuracy of over 85\% on a 11-class problem.

%\subsection{Authorship Obfuscation through Machine Translation}
The effect of automatic machine translation on text has been analyzed with focus on different aspects.
Detecting the tools which have been used to translate a text has been shown to be already possible for very small amounts of text \cite{Aharoni2014}.
Caliskan and Greenstadt  show that repeated translation of text does not obfuscate an author's individual writing style \cite{Caliskan2012}.
Instead, the translation itself adds even more information to the text, making it possible to determine which translation engine was used.
In their research, a text that has been translated two and three times is shown to still contain enough information to determine the authorship with an accuracy of over 90\%.
Similarly, it has been shown that the effectiveness of using machine-translation to obfuscate the original text and thus reduce the accuracy of authorship verification is limited \cite{Potthast2016}. 

%On the other hand, detecting the source language of a translated document has not been topic of much research.

%Being previously unexplored, we analyze the possibility of detecting the original language of a specific translated document rather than trying to identify the original author as a human.

\section{Methodology}

At a glance, we utilize textual features commonly used in stylometry to quantify the characteristics of translated English text originating from a specific source language.
These features are then used to train state-of-the-art machine-learning classifiers, which are finally used to predict the source language.

\subsection{Dataset}
\begin{figure}
  \centering
    \begin{tikzpicture}[
      >={Latex},
      my document/.style={documents}
    ]
    %\node (wiki) at (0, -2.5) {Wikipedia };
    %\node (lcc)  at (0, -1.5) {LCC };

    %\node[draw,align=center] (original) at (2, -2) {en\\es\\\ldots \\ nl\\zh};
    %\draw [shorten >= 2mm, ->] (wiki) to (original);
    %\draw [shorten >= 2mm, ->] (lcc) to (original);

    \node at (0,0) {\textbf{en\strut}};
    \node at (1,0) {\textbf{de\strut}};
    \node at (2,0) {\textbf{\ldots\strut}};
    \node at (3,0) {\textbf{nl\strut}};
    \node at (4,0) {\textbf{zh\strut}};

    \draw [line width=0.25mm] (-2mm, -5mm) to (2mm,-5mm);
    \draw [line width=0.25mm] (-2mm, -6mm) to (2mm,-6mm);
    \draw [line width=0.25mm] (-2mm, -7mm) to (2mm,-7mm);
    \draw [line width=0.25mm] (-2mm, -8mm) to (2mm,-8mm);

    \draw [line width=0.25mm] (8mm, -5mm) to (12mm,-5mm);
    \draw [line width=0.25mm] (8mm, -6mm) to (12mm,-6mm);
    \draw [line width=0.25mm] (8mm, -7mm) to (12mm,-7mm);
    \draw [line width=0.25mm] (8mm, -8mm) to (12mm,-8mm);

    \node at (20mm, -6.5mm) {\ldots};

    \draw [line width=0.25mm] (28mm, -5mm) to (32mm,-5mm);
    \draw [line width=0.25mm] (28mm, -6mm) to (32mm,-6mm);
    \draw [line width=0.25mm] (28mm, -7mm) to (32mm,-7mm);
    \draw [line width=0.25mm] (28mm, -8mm) to (32mm,-8mm);

    \draw [line width=0.25mm] (38mm, -5mm) to (42mm,-5mm);
    \draw [line width=0.25mm] (38mm, -6mm) to (42mm,-6mm);
    \draw [line width=0.25mm] (38mm, -7mm) to (42mm,-7mm);
    \draw [line width=0.25mm] (38mm, -8mm) to (42mm,-8mm);

    \draw [decorate, decoration={brace,mirror}] (-3mm, -5mm) -- (-3mm, -8mm) node [midway,xshift=-4mm] {\footnotesize{5k}};

    \node [draw, text width=7cm, align=right] at (2.5, -6.5mm) {\strut wiki};

    \node at (2,-11.5mm) {+};

    \draw [line width=0.25mm] (-2mm, -15mm) to (2mm,-15mm);
    \draw [line width=0.25mm] (-2mm, -16mm) to (2mm,-16mm);
    \draw [line width=0.25mm] (-2mm, -17mm) to (2mm,-17mm);
    \draw [line width=0.25mm] (-2mm, -18mm) to (2mm,-18mm);

    \draw [line width=0.25mm] (8mm, -15mm) to (12mm,-15mm);
    \draw [line width=0.25mm] (8mm, -16mm) to (12mm,-16mm);
    \draw [line width=0.25mm] (8mm, -17mm) to (12mm,-17mm);
    \draw [line width=0.25mm] (8mm, -18mm) to (12mm,-18mm);

    \node at (20mm, -16.5mm) {\ldots};

    \draw [line width=0.25mm] (28mm, -15mm) to (32mm,-15mm);
    \draw [line width=0.25mm] (28mm, -16mm) to (32mm,-16mm);
    \draw [line width=0.25mm] (28mm, -17mm) to (32mm,-17mm);
    \draw [line width=0.25mm] (28mm, -18mm) to (32mm,-18mm);

    \draw [line width=0.25mm] (38mm, -15mm) to (42mm,-15mm);
    \draw [line width=0.25mm] (38mm, -16mm) to (42mm,-16mm);
    \draw [line width=0.25mm] (38mm, -17mm) to (42mm,-17mm);
    \draw [line width=0.25mm] (38mm, -18mm) to (42mm,-18mm);

    \draw [decorate, decoration={brace,mirror}] (-3mm, -15mm) -- (-3mm, -18mm) node [midway,xshift=-4mm] {\footnotesize{5k}};
    \node [draw, text width=7cm, align=right] at (2.5, -16.5mm) {\strut LCC$^{1}$};
    \draw [->] (2,-20mm) to (2,-23mm);

    \draw [line width=0.25mm] (-2mm, -25mm) to (2mm,-25mm);
    \draw [line width=0.25mm] (-2mm, -26mm) to (2mm,-26mm);
    \draw [line width=0.25mm] (-2mm, -27mm) to (2mm,-27mm);
    \draw [line width=0.25mm] (-2mm, -28mm) to (2mm,-28mm);

    \draw [line width=0.25mm] (8mm, -25mm) to (12mm,-25mm);
    \draw [line width=0.25mm] (8mm, -26mm) to (12mm,-26mm);
    \draw [line width=0.25mm] (8mm, -27mm) to (12mm,-27mm);
    \draw [line width=0.25mm] (8mm, -28mm) to (12mm,-28mm);

    \node (original) at (20mm, -26.5mm) {\ldots};

    \draw [line width=0.25mm] (28mm, -25mm) to (32mm,-25mm);
    \draw [line width=0.25mm] (28mm, -26mm) to (32mm,-26mm);
    \draw [line width=0.25mm] (28mm, -27mm) to (32mm,-27mm);
    \draw [line width=0.25mm] (28mm, -28mm) to (32mm,-28mm);

    \draw [line width=0.25mm] (38mm, -25mm) to (42mm,-25mm);
    \draw [line width=0.25mm] (38mm, -26mm) to (42mm,-26mm);
    \draw [line width=0.25mm] (38mm, -27mm) to (42mm,-27mm);
    \draw [line width=0.25mm] (38mm, -28mm) to (42mm,-28mm);
    \draw [decorate, decoration={brace,mirror}] (-3mm, -25mm) -- (-3mm, -28mm) node [midway,xshift=-4mm] {\footnotesize{10k}};
    \node [draw, text width=7cm, align=right] at (2.5, -26.5mm) {\strut mixed};

    \node [draw, align=center] (google)    at (0,-4) {Google\strut};
    \node [draw, align=center] (microsoft) at (4,-4) {Microsoft\strut};

    \draw[shorten >= 3mm, shorten <= 6mm, ->] (original.south) to (microsoft.north);
    \draw[shorten >= 3mm, shorten <= 6mm, ->] (original.south) to (google.north);

    \node[draw, align=left] (tgoogle) at    (0,-6.2) {en = en {\footnotesize(untranslated)}\\es $\to$ en\\\ldots\\nl $\to$ en\\zh $\to$ en};
    \node[draw, align=left] (tmicrosoft) at (4,-6.2) {en = en {\footnotesize(untranslated)}\\es $\to$ en\\\ldots\\nl $\to$ en\\zh $\to$ en};
    
    \draw[shorten >= 1mm, shorten <= 1mm, ->] (google.south) to (tgoogle.north);
    \draw[shorten >= 1mm, shorten <= 1mm, ->] (microsoft.south) to (tmicrosoft.north);

    \node[docs] at (7mm, -85mm) {};
    \node[docs] at (6mm, -86mm) {\tiny XL};

    \node(dotsg) at (0mm, -86mm) {\ldots};

    \node[docs] at (-7mm, -85mm) {};
    \node[docs] at (-8mm, -86mm) {\tiny S};
    
    % Microsoft Documents
    \node[docs] at (47mm, -85mm) {};
    \node[docs] at (46mm, -86mm) {\tiny XL};

    \node(dotsm) at (40mm, -86mm) {\ldots};

    \node[docs] at (33mm, -85mm) {};
    \node[docs] at (32mm, -86mm) {\tiny S};

    \draw[shorten >= 3mm, shorten <= 3mm, ->] (tgoogle.south) to (dotsg.north);
    \draw[shorten >= 3mm, shorten <= 3mm, ->] (tmicrosoft.south) to (dotsm.north);

  \end{tikzpicture}
  \caption{Dataset Generation Workflow}
  \label{fig:workflow}
\end{figure}
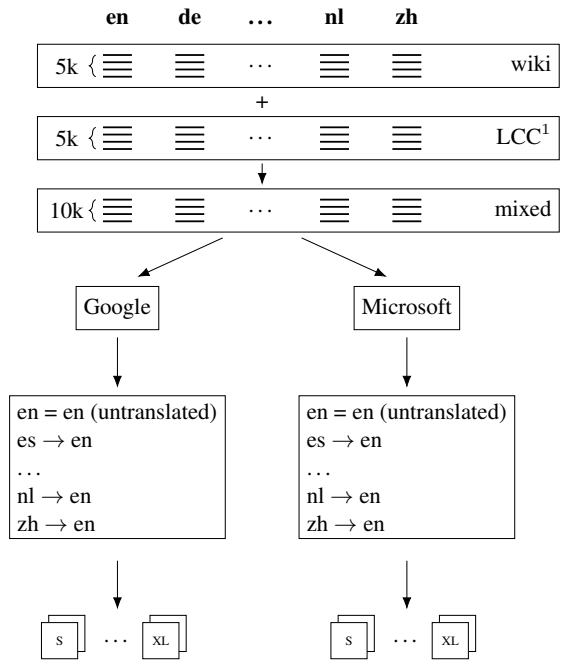

We used two different sources to construct a representative dataset containing German (de), English (en), Spanish (es), French (fr), Italian (it), Japanese (ja), Korean (ko), Dutch (nl), Turkish (tr) and Chinese (zh) text:
firstly, the Leipzig Corpus Collection (further called \textit{lcc}) \cite{Goldhahn} contains random sentences from news articles, web pages and Wikipedia articles.
Secondly, random Wikipedia articles were selected for each of the languages mentioned above, which will collectively be referred to as \textit{wiki}.
For each language from both \textit{lcc} and \textit{wiki}, 5,000 sentences were chosen randomly and translated to English using the services of Google\footnote{https://translate.google.com} and Microsoft\footnote{https://www.microsoft.com/en-us/translator/}.
Subsequently, 9 of the 10 languages are translated, whereas the original English sentences are not translated.
The dataset generation process is displayed on Fig.~\ref{fig:workflow}, where in the first step the sentences from both datasets are combined, yielding 10,000 sentences for each language.

Thereby, the translation itself was performed sentence-wise.
Both translation engines feature an automatic detection of the source language, however, since this information was known it was provided to them.
This implicitly enables our prediction model to detect whether a document has been machine-translated at all.
For all further investigations, the texts from the two services were kept separate, i.e., the computed models were tested on documents coming from the same translation service that was also used for training.
While the performance of cross-translation-service evaluations might be of interest at a first glance, this was omitted since it has already been shown that detecting the tool which was used to translate a text can be detected easily \cite{Aharoni2014}, i.e., in a realistic scenario one would probably use such an approach as a preliminary step to choose the correct model.
%\todo{Figure etwas genauer machen (siehe Schmierzettel ;)) - würde die Sprachen auch lowercase schreiben, also z.B. 'en' statt 'EN' (so hab ichs halt oft gesehen / PAN etc.)}
\subsubsection{Document Sampling}
To answer parts of RQ2, i.e., the influence of text length on the source language prediction quality, documents of various text sizes are required. 
To construct expressive documents, the translated sentences were combined randomly to create documents of various sizes, i.e., data sets consisting of short (S), medium (M), long (L) and very long (XL) documents, respectively\footnote{The sampling on a random basis was required since the \emph{lcc} dataset only contains random sentences, rather than cohesive documents.}.
The outcome of this step of the workflow is depicted in the bottom part of Fig. \ref{fig:workflow}, and details about the resulting datasets are listed in Table \ref{tab:documents}.
 
The choice for the document sizes is thereby loosely based on typical sizes of online texts according to social media statistics\footnote{e.g., https://blog.bufferapp.com/optimal-length-social-media}.
All documents created are unique with respect to their sentences contained, i.e., one sentence can only be part of one document per size class.
This limitation causes the data set with the longest documents (XL) to contain less documents than the other datasets, for which a maximum of 1,000 documents were created.

\begin{table}
  \caption{Data Sets with Varying Document Lengths}
  \centering
  \begin{tabular}{ l c c }
    \toprule
    length & no. of words & no. of docs \\
    \midrule
    short (S)  & 5--50 & 1,000 \\
    medium (M) & 51--200 & 1,000 \\
    long (L)  & 201--1,000 & 1,000 \\
    xlong (XL) & $\geq$ 1,001 & 312 (on avg) \\
    \bottomrule
  \end{tabular}
  \label{tab:documents}
\end{table}

\subsection{Features}
To distinguish the origin of the translated documents, several text features are used, which are listed in Table \ref{tab:features}.
These features were calculated by the Python library scikit-learn\footnote{http://scikit-learn.org/} as well as the language processing library spaCy\footnote{https://spacy.io/} for POS tagging.
Regarding the features used, the punctuation occurrence counts the amount of selected punctuation marks\footnote{\verb=:;.,?!"#$%&()[]{}*+-\/^_`|~=} in each document.
Features 2 and 3 count the occurrences of distinct words, separated by whitespace, whereby feature 3 disregards all English stop words\footnote{We rely on the list of stop words provided by scikit-learn}.
Features 4-23 count the occurrences of n-grams of various lengths of characters and POS-tags, respectively.
The n-gram-features denote the combination of all n-grams ranging from length 1 to 4.
The \textit{tf-idf} term denotes that the respective measure is normalized using term frequency-inverse document frequency metric provided by scikit-learn\footnote{http://scikit-learn.org/stable/modules/feature\_extraction.html\#tfidf-term-weighting}. 

Combinations of features were tested, but provided no additional accuracy over using single features.
This is displayed in Section~\ref{sec:results}.

\begin{table}
  \caption{List of Utilized Features}
  \centering
  \begin{tabular}{rl}
    \toprule
    id~~~ & feature name \\
    \midrule
   1~~~& punctuation occurrence \\
   2~~~& word occurrence \\
   3~~~& word occurrence, w/o stop words \\
 4-8~~~& character 1, 2, 3, 4, n-grams \\
9-13~~~& character 1, 2, 3, 4, n-grams w. tf-idf\\
14-18~~~& POS-tag 1, 2, 3, 4, n-grams \\
19-23~~~& POS-tag 1, 2, 3, 4, n-grams w. tf-idf\\
%\midrule
%25 & POS n-grams w. tfidf + char n-grams w. tfidf\\
%26 & POS 4-grams w. tfidf + char 4-grams w. tfidf\\
%27 & POS n-grams + char n-grams\\
%28 & POS 4-grams + char 4-grams\\
   \bottomrule
  \end{tabular}
  \label{tab:features}
\end{table}

%Except for the word count feature, which represents a single value per document, all other features consist of a vector of values. 
%To effectively compare these values to one another, different norms exist. 
%The most commonly used norms are l0 (no normalization), l1 (Manhattan distance) and l2 (Euclidean distance). 
%For each feature with multiple values, the best norm is determined by performing a 10-fold cross validation with a linear SVM classifier. 
%For all further experiments, each feature has been normalized with the best suited norm.

\subsection{Classifier}
In order to determine the classifier for this task, a preliminary evaluation has been conducted on the S-dataset, using all features described in Table~\ref{tab:features} individually.
Thereby, several popular classifiers were tested: Naive Bayes, K Nearest Neighbors (KNN), Support Vector Machines (SVM) with Linear and RBF Kernels, Random Forest and a Multilayer Perceptron (MLP). 
This preliminary step was required due to the infeasibility of testing all classifiers with all experimental runs described in Section~\ref{sec:results}.
%Many of these classifiers feature a number of parameters to tune, like the C-threshold for SVMs or the number of hidden layer nodes in neural networks.
%An exhaustive search for all optimal parameters is expensive, considering all features listed in Table~\ref{tab:features}.
To estimate the classifier performances, the default parameters given by the \textit{scikit-learn} library have been used.
The results of this preliminary run are displayed in Table~\ref{tab:classifiers}. 
We observe that the linear SVM outperformed all other classifiers, if only by a small margin.
The neural network (MLP) has a similar performance, at the cost of a much longer runtime.
Being in line with other research in this area \cite{Joachims1998,Koppel2005} and adhering to the results of this test, the linear SVM was used for further experiments. 

\begin{table}[t]
\caption{Accuracy of Different Classifiers on the Short Data Set. The feature ids correspond to the features listed in Table~\ref{tab:features}. The best performing feature (with id=13) corresponds to character n-grams with tf-idf normalization.}
  \centering
  \begin{tabular}{lccccc}
\toprule
& \multicolumn{2}{c}{ Microsoft} & &  \multicolumn{2}{c}{ Google}\\
\cmidrule{2-3} \cmidrule{5-6}
classifier & score & ~feature id~ & & score & ~feature id~ \\
\midrule
Naive Bayes & 0.39 & 12 & & 0.31 & 12 \\
KNN         & 0.30 & 11 & & 0.24 & 12 \\
SVM (linear) & \textbf{0.44} & 13 & & \textbf{0.34} & 13 \\
SVM (RBF)    & 0.19 & 14 & & 0.18 & 14 \\
Random Forest  & 0.28 &  3 & & 0.21 &  3 \\
MLP     & \textbf{0.44} & 13 & & \textbf{0.34} & 13 \\
\bottomrule 
\end{tabular}
\label{tab:classifiers}

\end{table}

\subsection{Experiment Setup}

As argued previously, the texts translated by the two translation services have not been mixed for all following calculations, i.e., we never test documents translated by Google on models that have only seen documents translated by Microsoft and vice versa. 

To answer both RQs, two separate experiments were conducted:
In the first setup, the influence of the number of possible source languages on the resulting accuracy is analyzed. 
%Therefore, only the \emph{short} dataset is used, which should provide a more difficult prediction.
At first, a classification process has been performed on the \textit{short} documents, including all source languages.
Subsequently, each combination of 2, 3 and 4 possible source languages has been selected, whereby each combination also contains original, untranslated English documents.
The restriction that the combinations must contain the English documents ensures that each combination also contains a non-translated set of documents.
This way, any difference in classifying translated vs. original text can be determined as well.

The second experiment subsequently analyzes the influence of the document length on the classification performance.
Here, all possible languages are taken into account, resulting in a 10-class classification problem.

%\subsection{Evaluations}

%For each combination, each feature listed in Table~\ref{tab:features} is evaluated iteratively, and the maximum accuracy is recorded.
%?? The arithmetic mean of all combinations with the same amount of languages is then used to represent this size.?? \todo{den Satz versteh ich nicht, sorry.. ;)}

%Regarding RQ2, i.e., determining the amount of text required for an accurate prediction, all data sets containing different document lengths have been evaluated by allowing all languages to be the source.
\section{Results}
\label{sec:results}

To quantify the individual expressiveness of each feature, we first conducted isolated evaluations for each feature using the S-dataset.
The corresponding results can be found in Table~\ref{tab:single-features}.
For readability reasons, only selected features are shown in the figure, with the best performing feature printed in bold face.
Then, the best performing features were selected, and combinations thereof were tested. 
Results of these experiments are displayed in Table~\ref{tab:multiple-features}, which displays the best performing feature set (character n-grams with tf-idf normalization, i.e., utilizing combined character 1-, 2-, 3- and 4-grams) including other well-performing combinations.
Comparing these results to the previous single-feature evaluation of Table \ref{tab:single-features}, it can be seen that combining features (sets) does not improve the classification performance.

\begin{table}
\caption{Accuracy of all Single Features}
\centering
\begin{tabular}{lccc ccc ccc ccc}
  \toprule
 feature id &    1 &    2 &    3 &    4 &    5 &    6 &    7 &    8 &    9 &   10 &   11 &   12 \\
  Microsoft & 0.21 & 0.36 & 0.33 & 0.25 & 0.35 & 0.39 & 0.39 & 0.41 & 0.29 & 0.36 & 0.40 & 0.41 \\ 
     Google & 0.15 & 0.28 & 0.25 & 0.21 & 0.28 & 0.29 & 0.30 & 0.33 & 0.23 & 0.29 & 0.30 & 0.31 \\  \midrule
 feature id &   13 &   14 &   15 &   16 &   17 &   18 &   19 &   20 &   21 &   22 &   23 & \\
  Microsoft & \textbf{0.44} & 0.20 & 0.27 & 0.25 & 0.22 & 0.27 & 0.20 & 0.27 & 0.25 & 0.21 & 0.24 & \\ 
     Google & \textbf{0.34} & 0.18 & 0.24 & 0.24 & 0.22 & 0.26 & 0.18 & 0.24 & 0.23 & 0.22 & 0.24 & \\ \bottomrule
\end{tabular}
\label{tab:single-features}

\end{table}

\begin{table}
  \caption{Accuracy of Best Feature and Selected Combinations}
  \centering
  \begin{tabular}{lcc}
    \toprule
    & \multicolumn{2}{c}{accuracy} \\ \cmidrule{2-3}
    features used & Microsoft & Google \\
    \midrule
    char-n (tfidf)                 & \textbf{0.44} & \textbf{0.34} \\
    char-n (tfidf) + POS-n (tfidf) & 0.25 & 0.24 \\
    char-n (tfidf) + char-n & \textbf{0.44} & \textbf{0.34} \\
    char-n (tfidf) + char-4 (tfidf) & 0.42 & 0.34 \\
    char-n (tfidf) + char-4 & 0.41 & 0.30 \\
    \bottomrule

  \end{tabular}
  \label{tab:multiple-features}
\end{table}
\begin{table}
  \centering
  \caption{Part of the Confusion Matrix for Short Documents Translated by Google}
  \begin{tabular}{lcccc}
    \toprule
    & ~~de~~ & ~~en~~  & ~~es~~ & ~~nl~~ \\
    \midrule
    de & 57 & 24  & 23 & 39\\
    en & 7  & 178 & 4  & 11 \\
    es & 28 & 7   & 86 & 17 \\
    %fr & 25 & 20 & 25 & 78 & 19 & 9 & 9 & 20 & 4 & 16\\
    %it & 21 & 21 & 23 & 30 & 74 & 10 & 17 & 18 & 4 & 13\\
    %ja & 14 & 17 & 11 & 16 & 13 & 121 & 28 & 10 & 5 & 14\\
    %ko & 15 & 12 & 16 & 13 & 9 & 25 & 99 & 13 & 10 & 18\\
    nl & 26 & 15 & 26 & 72 \\
    %tr & 13 & 17 & 11 & 12 & 8 & 11 & 20 & 9 & 134 & 12\\
    %zh & 5 & 5 & 9 & 10 & 6 & 8 & 11 & 7 & 3 & 138\\
    \bottomrule
  \end{tabular}
  \label{tab:confusion-matrix}
\end{table}

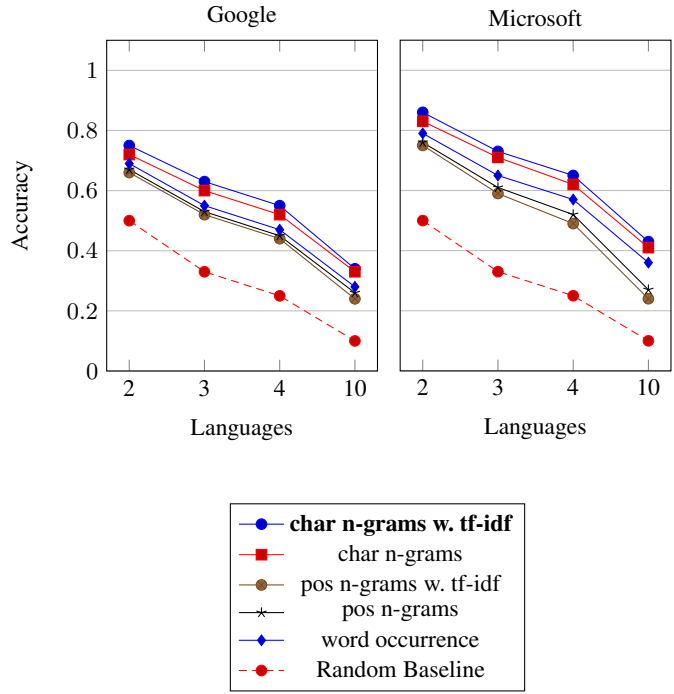
\begin{figure}
  \centering
  \begin{tikzpicture}
    \begin{groupplot}[
        group style={
          group size=2 by 1,
          horizontal sep=0.3cm,
          ylabels at=edge left,
          yticklabels at=edge left,
        },
        width=\textwidth,
        ymin=0.0,
        ymax=1.1,
        y=0.4cm/0.1,
        x=1cm/1,
        xtick=data,
        ytick style={draw=none},
        symbolic x coords={2,3,4,10},
        ytick={0,0.2,...,1.0},
        ylabel={Accuracy},
        x label style={at={(axis description cs:0.5,0)},anchor=north},
        y label style={at={(axis description cs:0.1,0.5)},anchor=south},
        title style={yshift=-1ex},
        %cycle list name=black white,
        legend style={
          font=\small,
          anchor=north,
          at={(1,-0.4)},
        },ymajorgrids=true
      ]
      \nextgroupplot[title=Google,xlabel=Languages]
      \addplot table[y=g-char_ngrams_tfidf] {ex1_features.csv}; \addlegendentry{\textbf{char n-grams w. tf-idf}};
      \addplot table[y=g-char_ngrams] {ex1_features.csv}; \addlegendentry{char n-grams};
      \addplot table[y=g-pos_ngrams_tfidf] {ex1_features.csv}; \addlegendentry{pos n-grams w. tf-idf};
      \addplot table[y=g-pos_ngrams] {ex1_features.csv}; \addlegendentry{pos n-grams};
      \addplot table[y=g-word_occurrence_with_stop_words] {ex1_features.csv}; \addlegendentry{word occurrence};
      \addplot table[y=Baseline] {ex1_features.csv};  \addlegendentry{Random Baseline};

      \nextgroupplot[title=Microsoft,xlabel=Languages]
      \addplot table[y=m-char_ngrams_tfidf] {ex1_features.csv}; ;
      \addplot table[y=m-char_ngrams] {ex1_features.csv}; ;
      \addplot table[y=m-pos_ngrams_tfidf] {ex1_features.csv}; ;
      \addplot table[y=m-pos_ngrams] {ex1_features.csv}; ;
      \addplot table[y=m-word_occurrence_with_stop_words] {ex1_features.csv}; ;
      \addplot table[y=Baseline] {ex1_features.csv};
    \end{groupplot}   
  \end{tikzpicture}
  \caption{Varying Number of Possible Languages using short document lengths}
  \label{fig:ex1_features}
\end{figure}

The overall performance on classifying the original language of a machine translated text is high. 
Fig.~\ref{fig:ex1_features} shows the classification performance with a varying amount of possible source languages.
It can be seen that for all text lengths tested, the prediction is well above the baseline, which represents random guessing out of the possible languages.
In case of only two possible source languages, text translated by Microsoft can be classified with an accuracy of over 0.85 when having only short documents as input.
The combination of character 1- to 4-grams with tf-idf normalization proves to be the best suiting feature set for both the Microsoft and Google translations.

\begin{figure}
  \centering
  \begin{tikzpicture}
    \begin{groupplot}[
        group style={
          group size=2 by 1,
          horizontal sep=0.3cm,
          ylabels at=edge left,
          yticklabels at=edge left,
        },
        width=\textwidth,
        ymin=0.0,
        ymax=1.1,
        y=0.4cm/0.1,
        x=1cm/1,
        xtick=data,
        ytick style={draw=none},
        symbolic x coords={S,M,L,XL},
        ytick={0,0.2,...,1.0},
        x label style={at={(axis description cs:0.5,0)},anchor=north},
        title style={yshift=-1ex},
        %cycle list name=black white,
        legend style={
          font=\small,
          anchor=north,
          at={(1,-0.4)},
        },ymajorgrids=true
      ]
      \nextgroupplot[title=Google,xlabel=Doc. Length]
      \addplot table[y=g-char_4grams_tfidf] {mixed.csv}; \addlegendentry{ \textbf{char 4-grams w. tf-idf}};
      \addplot table[y=g-char_4grams] {mixed.csv}; \addlegendentry{char 4-grams};
      \addplot table[y=g-pos_4grams_tfidf] {mixed.csv}; \addlegendentry{pos 4-grams w. tf-idf};
      \addplot table[y=g-pos_4grams] {mixed.csv}; \addlegendentry{pos 4-grams};
      \addplot table[y=g-word_occurrence_with_stop_words] {mixed.csv}; \addlegendentry{word occurrence};
      \addplot table[y=g-Baseline] {mixed.csv};  \addlegendentry{Random Baseline};

      \nextgroupplot[title=Microsoft,xlabel=Doc. Length]
      \addplot table[y=m-char_4grams_tfidf] {mixed.csv}; ;
      \addplot table[y=m-char_4grams] {mixed.csv}; ;
      \addplot table[y=m-pos_4grams_tfidf] {mixed.csv}; ;
      \addplot table[y=m-pos_4grams] {mixed.csv}; ;
      \addplot table[y=m-word_occurrence_with_stop_words] {mixed.csv}; ;
      \addplot table[y=m-Baseline] {mixed.csv};
    \end{groupplot}   
  \end{tikzpicture}
  \caption{Varying Document Lengths using 10 languages}
  \label{fig:ex2_features}
\end{figure}
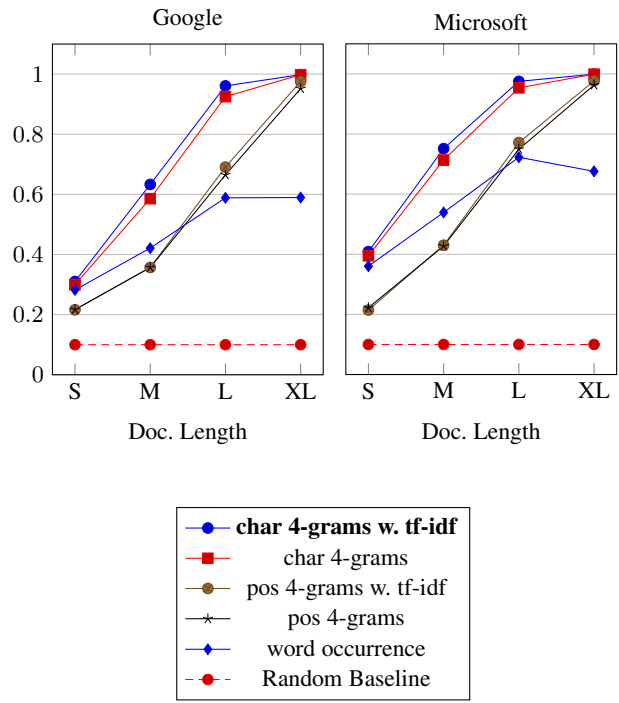

As expected and displayed in Figure~\ref{fig:ex2_features}, the quality of the predictions also increases substantially with the amount of text available.
For the XL-dataset, the accuracy of the classifier reaches values of over 0.99. 
This result suggests that if the amount of text translated by an engine is large enough, i.e., more than 1,000 words, its unique properties can be used to predict the source language nearly perfectly.


Both experiments show that character n-grams with tf-idf normalization achieve the best performance, regardless of the translation engine used.
%Detailed results of other features can be found in the supplemental material.
Another fact that can be extracted from the results is that the classification performs worse on text translated by Google rather than by Microsoft.
%In terms of authorship obfuscation, this might be interpreted as that Google is ``less worse'' than Microsoft in obfuscating authorship through automatic machine translation. 
It suggests that Google is currently able to transfer more complex linguistic structures onto its translation than Microsoft does.
Note that in general all results are based on the technology currently used by the two translation engines, which may change in the future.
While it can be assumed that the prediction accuracy decreases with improvements of the tools, future work should investigate this problem in detail.

%, which may be related to the fact that Google's translation engine has changed from statistical methods to neural networks\cite{Microsoft2016,Turovsky2016}.

Table~\ref{tab:confusion-matrix} shows an excerpt of the confusion matrix of short documents using the Google translation service.
Due to the origin of the languages it seems intuitive that some pairs (e.g., es--en) are easier distinguishable than others (e.g., nl--de), which is also reenforced by our experiments.
However and unfortunately, the SVM used for the classification provides no information about the relevance of the single character n-grams features, so we can't argue which parts of the translated text indicates the original language at this point.
Note that while other classifiers indeed can provide such information (e.g., decision trees or naive Bayes), their prediction accuracy is significantly worse, making statements about the indicators unreliable.

\bigskip

Answering RQ1, the experiments clearly show that it can be determined whether an English text is machine-translated, and that the source language can be predicted with high accuracy.
Moreover, the prediction quality increases significantly by restricting the amount of possible source languages as well as by providing longer texts (RQ2).

\section{Conclusion and Future Work}
In this paper, we have shown that, given an English text, it can be determined if it was translated by using stylometric features in combination with machine-learning.
Moreover, if translation was applied, the source language can be predicted at a high accuracy for regular length texts, and at a near-perfect accuracy when providing longer text samples. If the number of possible source languages is low, the prediction is reliable even for short length documents.
Possible extensions of this study include considering more different language classes (e.g., Arabic and African), analyzing other target languages for the translation or finding optimal parameters for different classifiers (e.g., Neural Networks).
However, while we showed that source language detection is possible at high accuracy in general, it remains unclear what the discriminating features (i.e., n-grams) are for the respective languages.
Using different classifiers which contain feature importances could be used to gain an intuitive understanding of the differences of specific language pairs.
Other approaches can include automatic measures like recursive feature elimination, which was omitted due to a limited timeframe.
Different kinds of features, such as grammar based features, may help to gain further understanding of the translation systems analyzed, which can subsequenly be leveraged to increase the performance of the model ever further. 

\bibliography{literature}
\bibliographystyle{plain}
\end{document}